\documentclass[a4paper]{article}
\usepackage{hyperref}
\usepackage{INTERSPEECH2018}
\usepackage{graphicx,balance,multicol}

\title{An Ensemble of Transfer, Semi-supervised and Supervised Learning Methods for Pathological Heart Sound Classification}

\name{\small Ahmed Imtiaz Humayun$^{1}$, Md. Tauhiduzzaman Khan$^{1}$, Shabnam Ghaffarzadegan$^{2}$, Zhe Feng$^{2}$ and Taufiq Hasan$^1$}
\address{\small 
  $^1$mHealth Lab, Dept. of Biomedical Engineering, Bangladesh University of Engineering and Technology (BUET), Bangladesh.\\
  $^2$Human Machine Interaction Group-2, Robert Bosch Research and Technology Center (RTC), Sunnyvale, CA.}
\email{taufiq@bme.buet.ac.bd, shabnam.ghaffarzadegan@us.bosch.com}

\begin{document}

\maketitle
\begin{abstract}
In this work, we propose an ensemble of classifiers to distinguish between various degrees of abnormalities of the heart using Phonocardiogram (PCG) signals acquired using digital stethoscopes in a clinical setting, for the INTERSPEECH 2018 Computational Paralinguistics (ComParE) Heart Beats Sub-Challenge. Our primary classification framework constitutes a convolutional neural network with 1D-CNN time-convolution (tConv) layers, which uses features transferred from a model trained on the 2016 Physionet Heart Sound Database. We also employ a Representation Learning (RL) approach to generate features in an unsupervised manner using Deep Recurrent Autoencoders and use Support Vector Machine (SVM) and Linear Discriminant Analysis (LDA) classifiers. Finally, we utilize an SVM classifier on a high-dimensional segment-level feature extracted using various functionals on short-term acoustic features, i.e., Low-Level Descriptors (LLD). An ensemble of the three different approaches provides a relative improvement of 11.13\% compared to our best single sub-system in terms of the Unweighted Average Recall (UAR) performance metric on the evaluation dataset.
%PCG signals are classified depending on the degree of abnormality (Normal, Mild, Severe).
\end{abstract}


\section{Introduction}
Cardiac auscultation is the most practiced non-invasive and cost-effective procedure for the early diagnosis of various heart diseases. Effective cardiac auscultation requires trained physicians, a resource which is limited especially in low-income countries of the world \cite{alam2010cardiac}. This lack of skilled doctors opens up opportunities for the development of machine learning based assistive technologies for point-of-care diagnosis of heart diseases. With the advent of smartphones and their increased computational capabilities, machine learning based automated heart sound classification systems implemented with a smart-phone attachable digital stethoscope in the point-of-care locations can be of significant impact for early diagnosis of cardiac diseases.
% , especially in low-income regions of the world.
%, particularly for countries that suffer from a shortage and geographic mal-distribution of skilled physicians.
%can accommodate self-assessment, 
%The practice of diagnosing heart sound abnormality from heart sounds require expert cognitive skills considering the wide range of variations possible. 
%Thus, 

Automated classification of the PCG, i.e., the heart sound, have been extensively studied and researched in the past few decades. Previous research on automatic classification of heart sounds can be broadly classified into two areas: (i) PCG segmentation, i.e., detection of the first and second heart sounds (S1 and S2), and (ii) detection of recordings as pathologic or physiologic. For the latter application, researchers in the past have utilized Artificial Neural Networks (ANN) \cite{uuguz2012ANN}, Support Vector Machines (SVM) \cite{gharehbaghi2015SVM} and Hidden Markov Models (HMM) \cite{saraccouglu2012HMM}. In, the 2016 Physionet/CinC Challenge was organized and an archive of $4430$ PCG recordings were released for binary classification of normal and abnormal heart sounds.
% Before the release of the 2016 PhysioNet/CinC Challenge dataset, the top heart sound classification methods employed Artificial Neural Networks (ANN) \cite{uuguz2012ANN}, Support Vector Machines (SVM) \cite{gharehbaghi2015SVM} and Hidden Markov Models (HMM) \cite{saraccouglu2012HMM}. The PhysioNet/CinC Challenge \cite{liu2016open} compiled an archive of $4430$ PCG recordings, which is the most extensive open-source heart sound dataset to date.
% split into an open training and hidden test dataset, which is the largest open-source heart sound dataset till date. 
This particular challenge encouraged new methods being utilized for this task. Notable features used for this dataset included, time, frequency and statistical features \cite{homsi2017}, Mel-frequency Cepstral Coefficients (MFCC) \cite{bobillo2016}, and Continuous Wavelet Transform (CWT). Most of the systems adopted the segmentation algorithm developed by Springer et al. \cite{springer2016segmentation}. Among the top scoring systems, Maknickas et al. \cite{maknickas2017} extracted Mel-frequency Spectral Coefficients (MFSC) from unsegmented signals and used a 2D CNN. 
%for classification which achieved an accuracy of 0.8415 on the hidden test set. 
%with 0.855 accuracy. 
Plesinger et al. \cite{plesinger2017} proposed a novel segmentation method, a histogram based feature selection method and parameterized sigmoid functions per feature, to discriminate between classes. Various machine learning algorithms including SVM \cite{whitaker2017}, k-Nearest Neighbor (k-NN) \cite{bobillo2016}, Multilayer Perceptron (MLP) \cite{kay2017,zabihi2016}, Random Forest \cite{homsi2017}, 1D \cite{potes2016ensemble} and 2D CNNs \cite{maknickas2017}, and Recurrent Neural Network (RNN) \cite{yang2016classification} were employed in the challenge. A good number of submissions used an ensemble of classifiers with a voting algorithm \cite{homsi2017,kay2017,zabihi2016,potes2016ensemble}. The best performing system was presented by Potes et al. \cite{potes2016ensemble} that combined a 1D-CNN model with an Adaboost-Abstain classifier using a threshold based voting algorithm.
%that achieved an overall accuracy of 86.02\%. This work 
% A noticeable find is that most of the submissions took a naive approach of using a weighted voting between multiple classifiers. The features used were not always discriminant between classes as reported in \cite{plesinger2017}. 

%% Brought some changes here, please review - Imtiaz
In audio signal processing, filter-banks are commonly employed as a standard pre-processing step during feature extraction. This was done in \cite{potes2016ensemble} before the 1D-CNN model.
We propose a CNN based Finite Impulse Response (FIR) filter-bank front-end, that automatically learns frequency characteristics of the FIR filterbank utilizing time-convolution (tConv) layers. The INTERSPEECH ComParE Heart Sound Shenzhen (HSS) Dataset is a relatively smaller corpus, with three class labels according to the degree of the disease; while the Physionet Heart Sounds Dataset has binary annotations. We train our model on the Physionet Challenge Dataset and transfer the learned weights for the three class classification task. We also avail unsupervised/semi-supervised learning to find latent representations of PCG.

% The pathological or physiological significance of heart sounds frequency components are not well defined like other biosignals, i.e. alpha, beta and gamma bands of the Electroencephalogram (EEG). Pathological information can also be present in the temporal coherence of cardiac cycles. 
% However, no particular physiological significance of the filter-bank structure and their cutoff frequency definitions were presented.
% In our work, we propose a CNN based Finite Impulse Response (FIR) filter-bank front-end, that automatically learns the frequency characteristics of the FIR filters utilizing a time-convolution (tConv) layer. 
% However, as the Physionet challenge only dealt with normal/abnormal heart sounds while the ComParE 2018 challenge aims to classify the degree of abnormality (Normal, Mild, Severe), we utilized a transfer learning scheme to utilize the parameters from the model trained on Physionet and apply them on the current task.
%% copied from EMBC 2018 paper. END

\section{Data Preparation}
\subsection{Datasets}
\subsubsection{The INTERSPEECH 2018 ComParE HSS Dataset}
The INTERSPEECH 2018 ComParE Challenge \cite{schuller2018interspeech} released the Heart Sounds Shenzhen PCG signal corpus containing $845$ recordings from $170$ different subjects. The recordings were collected from patients with coronary heart disease, arrhythmia, valvular heart disease, congenital heart disease, etc. The PCG recordings are sampled at $4$ KHz and annotated with three class labels: (i) \emph{Normal}, (ii) \emph{Mild}, and (iii) \emph{Moderate/Severe} (heart disease). 
% Annotations were confirmed with echo-cardiogram. Data was partitioned into train/dev/test set for the competition. Details are discussed on the challenge paper \cite{schuller2018interspeech}.
\subsubsection{PhysioNet/CinC Challenge Dataset}
The 2016 PhysioNet/CinC Challenge dataset \cite{liu2016open} contains PCG recordings from seven different research groups. The training data contains $3153$ heart sound recordings collected from $764$ patients with a total number of $84,425$ annotated cardiac cycles ranging from $35$ to $159$ bpm. Cardiac Anomalies range from coronary heart disease, arrhythmia, valvular stenosis/regurgitation, etc. The dataset has $2488$ and $665$ PCG signals annotated as \emph{Normal} and \emph{Abnormal}, respectively. The Aristotle University of Thessaloniki heart sounds database (AUTHHSDB) \cite{papadaniil2014efficient}, a subset of the Physionet corpus (training-c), contains additional metadata based on the severity of the heart diseases. The recordings are sampled at 2000 Hz.
\subsection{Data Imbalance Problem}
The INTERSPEECH ComParE HSS Dataset suffers from significant class imbalance in its training set, which could introduce performance reduction for both classical machine learning and deep learning based classifiers. The training set is divided in a ratio of 16.7/55.0/28.3 percent between the \emph{Normal}/\emph{Mild}/\emph{Severe} classes, with more than half of the training data comprising of PCG signals annotated as \emph{Mild}". The result of the imbalance was evident in our recall metrics which are discussed later in Sec. \ref{disc}.
% \subsection{Oversampling/undersampling}
\subsection{Fused Training Sets} \label{database}
\begin{figure}[tb]
\includegraphics[width=\linewidth]{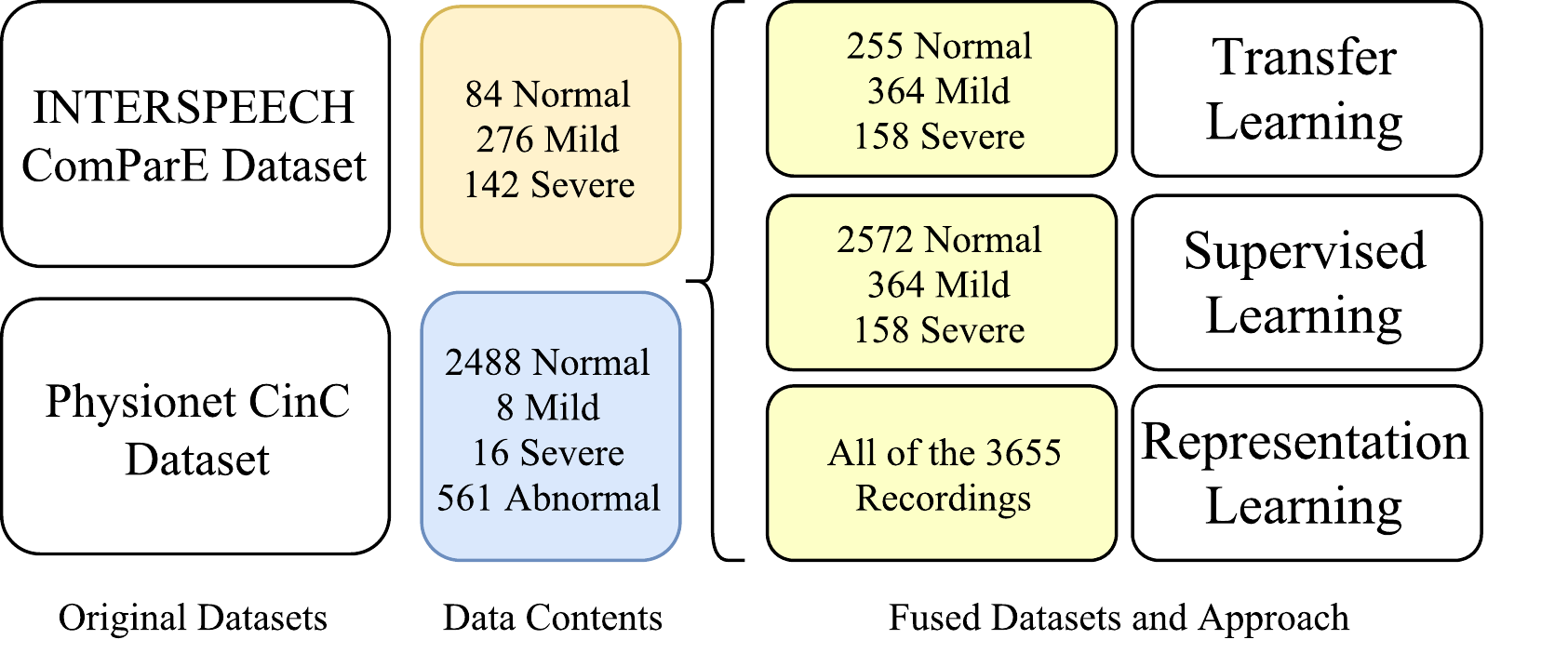}
\centering
\caption{Dataset preparation for transfer Learning, supervised Learning and representation learning using Physionet and ComParE corpus.}
\label{foldsplit}
% \vspace{-3mm}
\end{figure}

To cope with the class imbalance and increase the volume of the training data, we created $3$ new fused training corpora out of the INTERSPEECH ComParE HSS Dataset and the Physionet/CinC Challenge Dataset training partitions. The AUTHHSDB (training-c) partition of the dataset was relabeled using the metadata files provided to have $7$ \emph{Normal}, $8$ \emph{Mild} and $16$ \emph{Severe} annotated recordings. The dataset distributions are depicted in Fig. \ref{foldsplit}. The fused datasets prepared for Transfer Learning (TL), Supervised Learning (SL) and Representation Learning (RL) will be referred to as TL-Data, SL-Data and RL-Data respectively.
%\section{Transferring Learned Parameters from Physionet Model}
\section{Proposed Transfer Learning Framework}
\subsection{1D-CNN Model for Abnormal Heart Sound Detection} \label{potes}
The Physionet/CinC Challenge PCG database is a larger corpus with Normal and Abnormal labels designed for a binary classification task. We propose a 1D-CNN Neural Network improving the top scoring model \cite{potes2016ensemble} of the Physionet/CinC 2016 challenge. First, the signal is re-sampled to 1000 Hz (after an anti-aliasing filter) and decomposed into four frequency bands ($25-45$, $45-80$, $80-200$, $200-500$ Hz). Next, spikes in the recordings are removed \cite{schmidt2010segmentation} and PCG segmentation is performed to extract cardiac cycles \cite{springer2016segmentation}. Taking into account the longest cardiac cycle in the corpus, each cardiac cycle is zero padded to be $2.5$s in length. Four different frequency bands of extracted from each cardiac cycle are fed into four different input branches of the 1D-CNN architecture. Each branch has two convolutional layers of kernel size $5$, followed by a Rectified Linear Unit (ReLU) activation and a max-pooling of $2$. The first convolutional layer has $8$ filters while the second has $4$. The outputs of the four branches are fed to an MLP network after being flattened and concatenated. The MLP network has a hidden layer of $20$ neurons with ReLU activation and two output neurons with softmax activation. The resulting model provides predictions on every heart sound segment (cardiac cycle), which are averaged over the entire recording and rounded for inference.
\begin{figure*}[t]
\includegraphics[width=\linewidth]{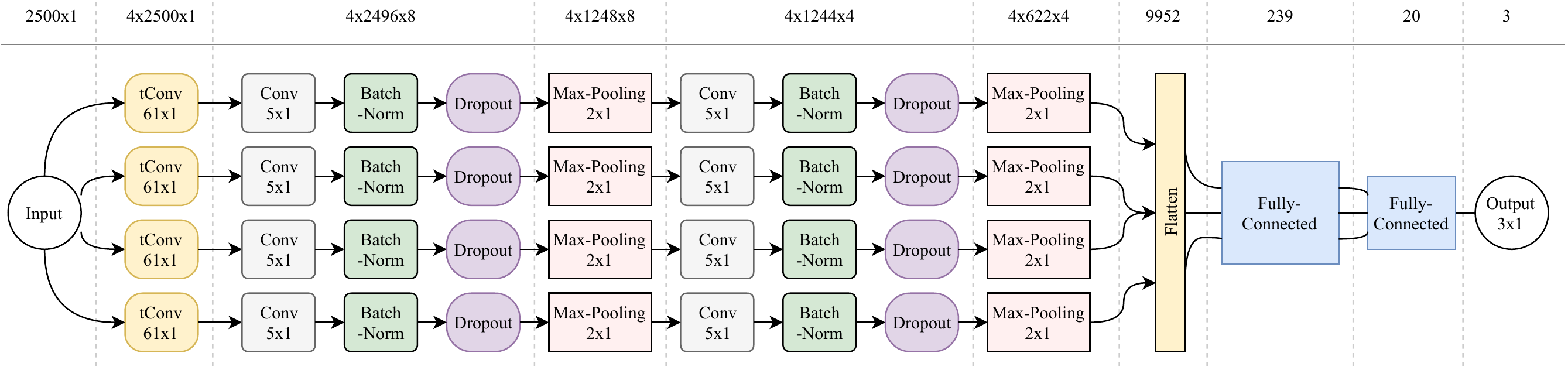}
\centering
\caption{Proposed architecture incorporating tConv layers for Transfer Learning.}
\label{bottle}
% \vspace{-3mm}
\end{figure*}
\subsection{Filter-bank Learning using Time-Convolutional (tConv) Layers}
For a causal discrete-time FIR filter of order $N$ with filter coefficients $b_{0}, b_{1}, \dots b_{N}$, the output signal samples $y[n]$ is obtained by a weighted sum of the most recent samples of the input signal $x[n]$. This can be expressed as: 
\begin{eqnarray}
%\begin{split}
y[n] &=& b_{0}x[n]+b_{1}x[n-1]+.....+b_{N}x[n-N] \nonumber\\
 &=& \sum_{i=0}^{N}b_{i}x[n-i].
%\end{split}
\end{eqnarray}
%CNNs are inspired by the neural interconnectivity present in the visual cortex of animals. They have spatially contiguous receptive fields. 
% Through a local connectivity pattern of neurons between adjacent layers, a 1D-CNN performs cross-correlation between its input and its kernel.
A 1D-CNN performs cross-correlation between its input and its kernel using a spatially contiguous receptive field of kernel neurons. The output of a convolutional layer, with a kernel of odd length $N+1$, can be expressed as: 
%is also a weighted sum of the input:
\begin{align}\label{conveqn}
%\begin{split}
\small
y[n] & = b_{0}x[n+\tfrac{N}{2}]+b_{1}x[n+\tfrac{N}{2}-1]+....+b_{\tfrac{N}{2}}x[n]+.... \nonumber\\
& +b_{N-1}x[n-\tfrac{N}{2}+1]+b_{N}x[n-\tfrac{N}{2}] \nonumber\\
& = \sum_{i=0}^{N}b_{i}\hspace{0.5mm}x[n+\tfrac{N}{2}-i]
%\end{split}
\end{align}
where $b_{0}, b_{1}, ... b_{N}$ are the kernel weights. Considering a causal system the output of the convolutional layer becomes:
\begin{equation}
y[n - \tfrac{N}{2}] = \sigma\left(\beta +\sum_{i=0}^{N}b_{i}x[n-i]\right)
\end{equation}
where $\sigma(\cdot)$ is the activation function and $\beta$ is the bias term. Therefore, a 1D convolutional layer with linear activation and zero bias, acts as an FIR filter with an added delay of $N/2$ \cite{matei2006CNNFIR}. We denote such layers as time-convolutional (tConv) layers \cite{sainath2015google}. Naturally, the kernels of these layers (similar to filter-bank coefficients) can be updated with Stochastic Gradient Descent (SGD). These layers therefore replace the static filters that decompose the pre-processed signal into four bands (Sec. \ref{potes}). We use a special variant of the tConv layer that learns coefficients with a linear phase (LP) response.

\subsection{Transfer Learning from Physionet Model}
Our proposed tConv Neural Network is trained on the Physionet CinC Challenge Dataset with four-fold in house cross validation \cite{ahmed2018}. The model achieves a mean cross-validation accuracy of $87.10\%$ and Recall of $90.91\%$. The weights up-to the flatten layer are transferred \cite{yosinski2014transferable} to a new convolutional neural network architecture with a fully connected layer with two hidden layers of 239 and 20 neurons and 3 output neurons for \emph{Normal}, \emph{Mild} and \emph{Severe} classes (Fig. \ref{bottle}). The model weights are fine-tuned on TL-Data. TL-Data comprises of all of the samples from the INTERSPEECH ComParE Dataset and the \emph{Normal} signals from the Physionet in house validation fold, from which the trained weights are transferred. We chose the weights of a model trained on Fold 1 for better per cardiac cycle validation accuracy. The cross-entropy loss is optimized with a stochastic gradient descent optimizer with a learning rate of $4.5*10^{-05}$. Dropout of $0.5$ is applied to all of the layers except for the output layer. The model hyperparameters were \emph{not} optimized while fine-tuning with TL-Data. The cost function was weighted to account for the class imbalance.

\section{Representation Learning (RL) with Recurrent Autoencoders}

\begin{figure}[h]
\includegraphics[width=\linewidth]{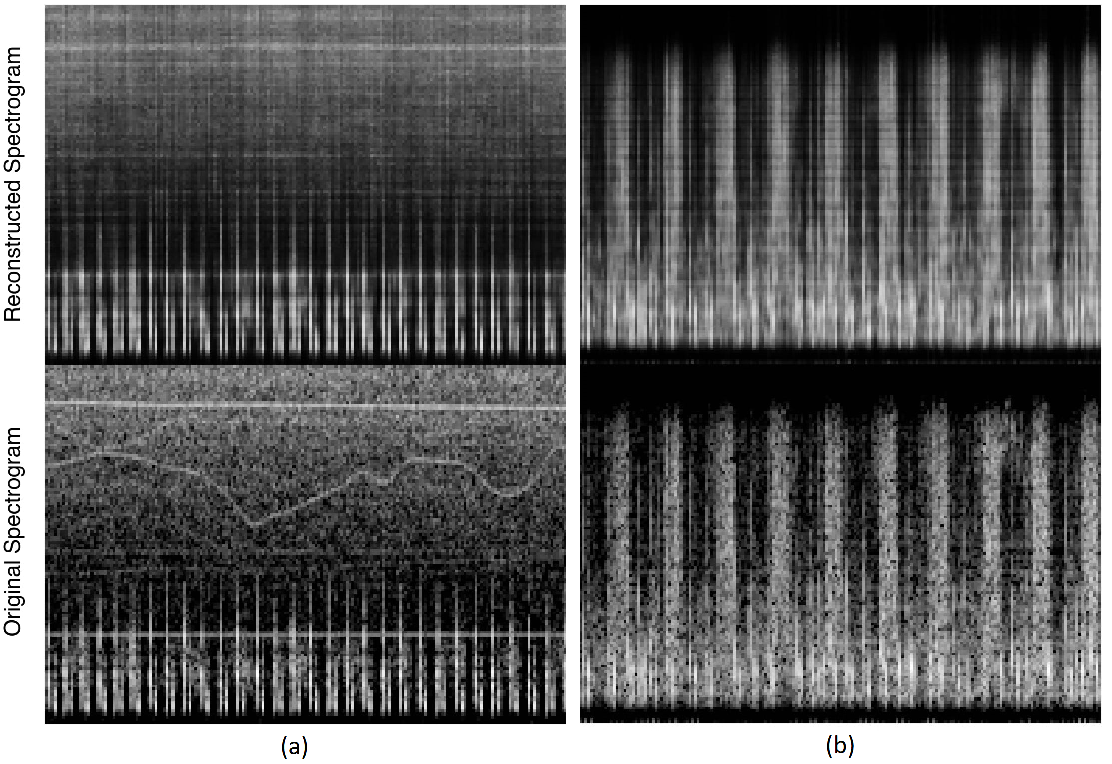}
\centering
\caption{Reconstructed Mel-spectrogram of recording thresholded to reduce background noise a) below -30 dB b) below -45 dB}
\label{recon}
% \vspace{-3mm}
\end{figure}

Representation learning is particularly of interest when a large amount of unlabeled data is available compared to a smaller labeled dataset. Considering the two corpora at hand, we approach the problem from a semi-supervised representation learning perspective to train recurrent sequence to sequence autoencoders \cite{sutskever2014sequence} on unlabeled RL-Data (Sec. \ref{database}) and then use lower dimensional representations of SL-Data to train classifiers. Sequence-to-sequence learning is about translating sequences from one domain to another. Unsupervised Sequence-to-sequence representation learning was popularized in the use of machine translation \cite{bahdanau2014neural}. It has also been employed for audio classification with success \cite{amiriparian2017sequence}. It offers the chance of resolving the overfitting problem experienced when training an end to end deep learning model.

First, mel-spectrogram of $126$ bands are extracted with a window size of $320$ms with $50$\% overlap. The raw audio files are clipped to 30 seconds in length. To reduce background noise, the spectrogram is thresholded below $-30$,$-45$,$-60$ and $-75$ dB. This results in four different spectrograms. The model is trained on all four of these separately, which results in four different feature sets. Both the encoder and decoder Recurrent Neural Network had 2 hidden layers with 256 Gated Recurrent Units each. The final hidden states of all the GRUs are concatenated into a 1024 dimensional feature vector. Fig. \ref{recon} portrays the reconstructed outputs for mel-spectrograms clipped below $-30$ dB and $-45$ dB. Four different feature vectors for the four different spectrograms are also concatenated to form fused features. 
Feature representations of SL-Data were used to train classifiers. The model is deployed and trained using the {\scriptsize AU}{D}{\scriptsize EEP} toolkit \cite{freitag2017audeep}.

% Classifier training
\section{Supervised Learning with Segment-level Features}
%\section{Classification using Segment-level Features}
\subsection{ComParE Acoustic Feature Set}
In this sub-system, we utilize the acoustic feature set described in \cite{weninger2013acoustics}.
This feature set contains $6373$ static features resulting from the computation of various functionals over LLD parameters \cite{schuller2018interspeech}. The LLD parameters and functionals utilized are described in \cite{weninger2013acoustics}. 
%Table \ref{opensmile_lld} and \ref{opensmile_functionals}, respectively. 
The features are extracted using the openSMILE toolkit \cite{eyben2010opensmile}.

\subsection{Classifiers}
% we may have several other classifiers later on. 
We have implemented several machine learning algorithms for heart sound classification from the ComParE Acoustic feature set. The evaluated classifiers include:
%Gaussian Mixture Model (GMM), 
Support Vector Machine (SVM), Linear Discriminant Analysis (LDA), and Multi-Layer Perceptron (MLP). SVM classifier with complexity C$=10^{-4}$ and tolerance L$=0.3$ outperformed the other classifiers.
%Naive Bayes (NB), Decision Tree (DT), k-Nearest Neighbor (kNN) and Random Forest (RF). 
% The classifier parameters as used in this work are summarized in Table \ref{classifier_params}.

% \begin{table}[t]
% \centering
% \caption{Implemented Classifiers and Training Parameters}
% \label{classifier_params}
% \begin{tabular}{c|l}
% \hline
% \hline
% Classifiers& \multicolumn{1}{c}{Parameters}\\ \hline
% % \begin{tabular}[c]{@{}c@{}}Gaussian Mixture Model\\ (GMM)\end{tabular} & \begin{tabular}[c]{@{}l@{}}No. of mixture components: 16\\ No. of EM iterations: 50\end{tabular}\\ 
% % \hline 
% \begin{tabular}[c]{@{}c@{}}Linear Discriminant Analysis\\ (LDA)\end{tabular}  & 
% \begin{tabular}[c]{@{}l@{}}Kernel: Linear\\ Optimization: Bayesian\end{tabular}\\ 
% \hline 
% \begin{tabular}[c]{@{}c@{}}Naive Bayes (NB)\end{tabular}            & \begin{tabular}[c]{@{}l@{}}Kernel: Normal Distribution \\ Max objective evaluation: 30\end{tabular} \\
% \hline 
% \begin{tabular}[c]{@{}c@{}}Decision Tree (DT)\end{tabular} & Max grid division: 10\\
% \hline
% \begin{tabular}[c]{@{}c@{}}k-Nearest Neighbour (kNN)\end{tabular}& 
% \begin{tabular}[c]{@{}l@{}}Number of neighbors: 5\\ Neighbour search method: kd-tree\end{tabular}\\ 
% \hline 
% \begin{tabular}[c]{@{}c@{}}Random Forest (RF)\end{tabular}& \begin{tabular}[c]{@{}l@{}}No. of Bags/tree: 76\\ Tree method: Classification\end{tabular}\\
% \hline
% \hline
% \end{tabular}
% \vspace{-2mm}
% \end{table}

\section{Experimental Evaluation and Results}\label{results}
% \subsection{Metrics}
The evaluation metric for the INTERSPEECH ComParE Challenge is Unweighted Average Recall (UAR) since the datasets are class unbalanced. We also monitor classwise recall and accuracy for evaluation of model performance. Performance metrics on both the development and test set are listed on Table \ref{metrics} with the training datasets mentioned. The Comp-SVM model, evaluated on the ComParE test set, acquired 45.9\% UAR and 51.5\% overall accuracy. Our transfer learning based model with a variant of our proposed tConv layer acquired improved performance compared to the end to end deep learning baseline (END2YOU). Training on a larger corpus has provided an improved performance on the development set using representation learning with significantly reduced performance on the test set. Comp-SVM, RL-SVM and LP-tConv models are ensembled using a majority voting algorithm. It yields UAR of 57.92\% on the development set, and UAR of 39.2\% on the test set. To improve the \emph{Normal} hit rate a hierarchical decision system is implemented where an LP-tConv network trained on Physionet/Cinc Database is first used for binary classification between \emph{Normal} and \emph{Abnormal} recordings. Following that, an ensemble of Comp-SVM, RL-SVM and LP-tConv is used to classify between \emph{Mild} and \emph{Severe} classes. The hierarchical model has acquired a dev set UAR of 57.93\% and test set UAR of 42.1\%. 
%Thus, according to our experiments, the results on the development set have not generalized well on the test set.

\begin{table}[t]
\centering
\caption{Performance evaluation of proposed methods compared to the official baseline systems.}
\label{metrics}
\resizebox{\linewidth}{!}{%
\begin{tabular}{|c|c|c|c|c|c|c|}
\hline
\hline
\multicolumn{7}{|c|}{\emph{Baseline Systems}}\\
\hline
\hline
Model Name & Dataset & Features & Classifiers & UAR (\%) dev  & Acc. (\%) dev & UAR (\%) test\\ \hline
OPENSMILE\cite{schuller2018interspeech}                                            & \begin{tabular}[c]{@{}c@{}}INTERSPEECH\\ ComParE HSS\end{tabular} & \begin{tabular}[c]{@{}c@{}}ComParE\\Feature set\end{tabular}                  & SVM         & 50.3 & 52.2 & 46.4     \\ \hline
AUDEEP\cite{schuller2018interspeech}                                               & \begin{tabular}[c]{@{}c@{}}INTERSPEECH\\ ComParE HSS\end{tabular} & \begin{tabular}[c]{@{}c@{}}Fused\\ 
Autoencoder\\ Features\end{tabular}        & SVM         & 38.6 & - & 47.9        \\ \hline
END2YOU\cite{schuller2018interspeech}                                              & \begin{tabular}[c]{@{}c@{}}INTERSPEECH\\ ComParE HSS\end{tabular} & CNN                                                                            & LSTM        & 41.2 & - & 37.7        \\ \hline
\multicolumn{4}{|c|}{Fusion of best 2 systems \cite{schuller2018interspeech}} & - & - & 56.2 \\\hline\hline
\multicolumn{7}{|c|}{\emph{Proposed Systems}}\\
\hline
\hline
Model Name & Dataset & Features & Classifiers & UAR (\%) dev  & Acc. (\%) dev & UAR (\%) test\\ \hline
ComP-SVM                                             & \begin{tabular}[c]{@{}c@{}}INTERSPEECH\\ ComParE HSS\end{tabular} & \begin{tabular}[c]{@{}c@{}}ComParE\\ Feature set\end{tabular}                  & SVM         & 52.1 & 53.9 & 45.9    \\ \hline
RL-SVM                                               & \begin{tabular}[c]{@{}c@{}}RL-Data\\ SL-Data\end{tabular}         & \begin{tabular}[c]{@{}c@{}}$-60$ dB\\ Autoencoder\\ Features\end{tabular}        & SVM         & 42.9 & 48.9 & -    \\ \hline
RL-LDA                                               & \begin{tabular}[c]{@{}c@{}}RL-Data\\ SL-Data\end{tabular}         & \begin{tabular}[c]{@{}c@{}}$-60$ \& $-75$dB\\ Autoencoder\\ Features\end{tabular} & LDA         & 51.4 & 54.4  & 34.4   \\ \hline

LP-tConv                                             & TL-Data                                                           & \begin{tabular}[c]{@{}c@{}}tConv\\ CNN\end{tabular}                            & MLP         & 44.6 & 56.1 & 39.5    \\ \hline\hline
\multicolumn{7}{|c|}{\emph{System Ensembles}}\\
\hline
\hline
%\begin{tabular}[c]{@{}c@{}}Ensemble\\(Majority Vote)\end{tabular}                                            & -                                                          & -                            & -         & 57.9 & 63.9     \\ \hline
\multicolumn{4}{|c|}{Ensemble System Name} & UAR (\%) dev  & Acc. (\%) dev & UAR (\%) test\\ \hline
\multicolumn{4}{|c|}{Fusion of Comp-SVM, RL-SVM and LP-tConv models} & 57.92 & 63.9 & 39.3 \\\hline
\multicolumn{4}{|c|}{Hierarchical with Fusion} & 57.93 & 64.2 & 42.1 \\\hline
\end{tabular}%
}
\end{table}

\begin{figure}[t]
\includegraphics[width=\linewidth,trim={1cm 3.2cm 2cm 1cm}]{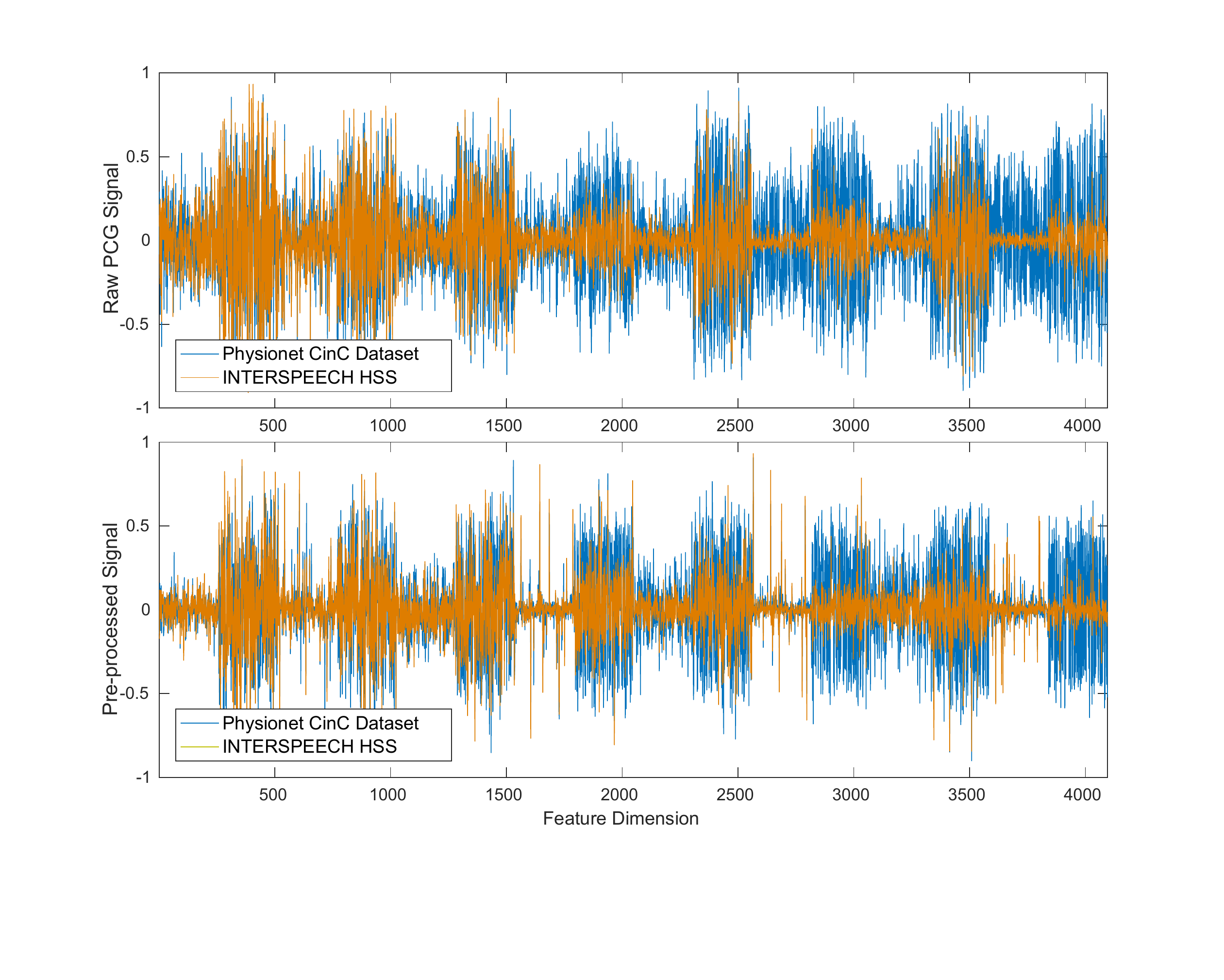}
\centering
\caption{Mean values of the $4096$ features learned from the 4 mel-spectrograms by the RNN-Autoencoders.}
\label{encoding}
% \vspace{-3mm}
\end{figure}

% \subsection{Classifier}
% Using the Basic Machine Learning as described in Table. \ref{classifier_params}, the accuracy, precision, and recall achieved is listed in the Table. \ref{initial_res}.
% \begin{table}[ht]
% \centering
% \caption{Performance of Different Classifiers}
% \label{initial_res}
% \begin{tabular}{l|l|l|l}
% \hline
% \textbf{Classifier} & \textbf{Acc} & \textbf{Recall} & \textbf{Precision} \\ \hline
% \textbf{SVM}        & 0.7019       & 0.3739          & 0.5749             \\ \hline
% \textbf{LDA}        & 0.9852       & 0.9864          & 0.963              \\ \hline
% \textbf{RF}         & 0.7296       & 0.4274          & 0.6141             \\ \hline
% \textbf{kNN}        & 0.6074       & 0.3358          & 0.336              \\ \hline
% \textbf{DT}         & 1            & 1               & 1                  \\ \hline
% \textbf{NB}         & 0.981481     & 0.954932        & 0.977179           \\ \hline
% \end{tabular}
% \end{table}


\section{Discussion}\label{disc}

\begin{figure}[tb]
\includegraphics[width=\linewidth,trim={1cm 1cm 1 1cm}]{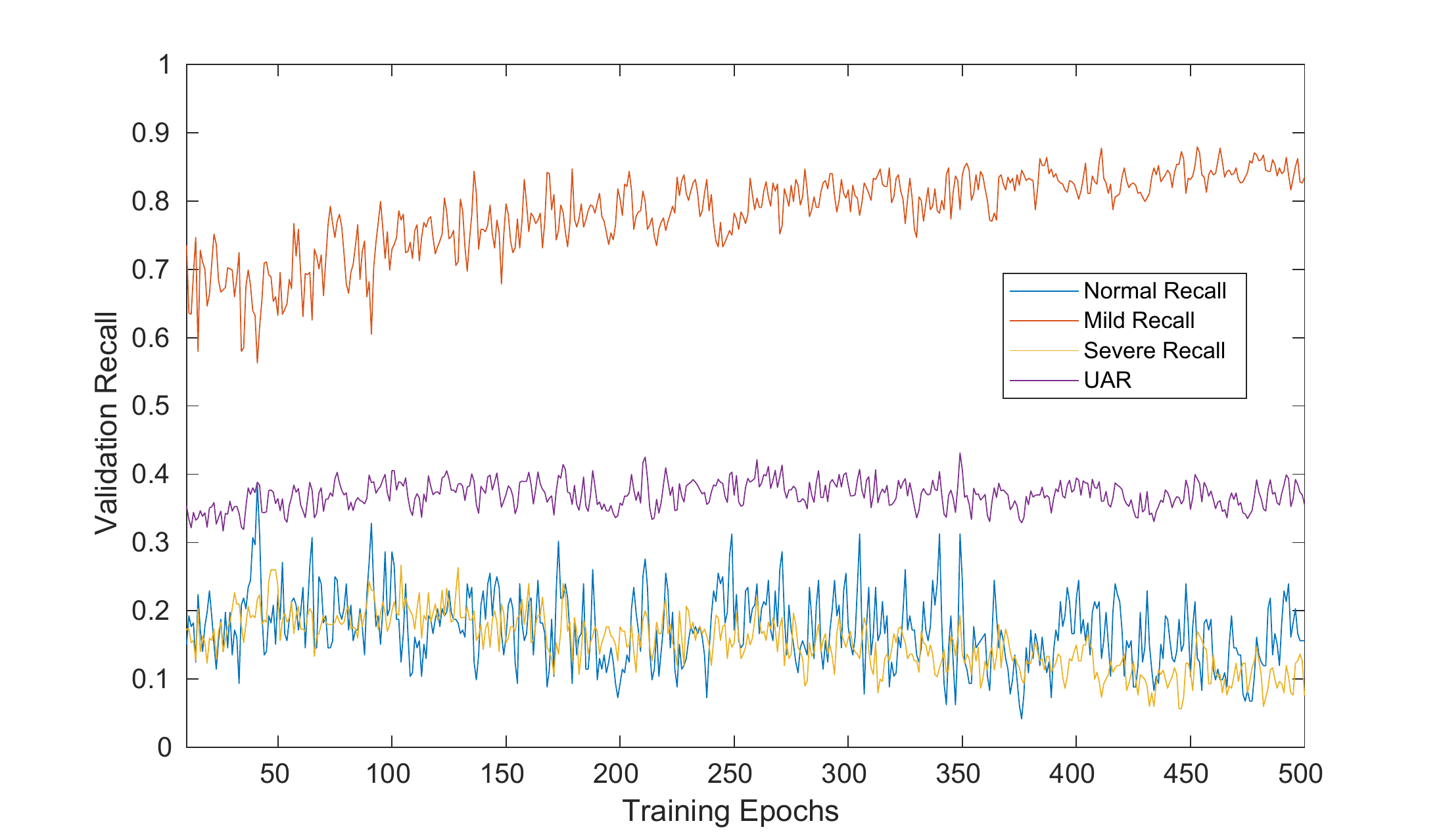}
\centering
\caption{Recall scores obtained on the validation data after each training epoch. A steady increase in the mild recall is visible while recall for the other classes are steadily decreasing.}
\label{recall}
% \vspace{-3mm}
\end{figure}
Our proposed end to end LP-tConv model superseded the test set metric for the standalone baseline end to end model (END2YOU). Other proposed systems failed to beat the baseline systems test set UAR while it outperformed the development set UAR. This could indicate overfitting on the development set. On the other hand, for the baseline systems a tendency of overfitting on the test set was visible. This is because the individual approach/hyperparameters performing best on the test set has been chosen as baselines \cite{schuller2018interspeech}. A generalized feature-classifier system should yield similar UAR on both development and test dataset if the development and test data distributions are consistent. This was noticeable only for the openSMILE features with an SVM classifier. 

More interesting insights were revealed during the training of the recurrent autoencoders. The lower dimensional representations learned were different for the Physionet CinC Challenge database and the INTERSPEECH ComParE HSS database. The RL model was trained on both RL-data and the INTERSPEECH HSS database. Fig. \ref{encoding} shows the mean of the concatenated (fused) representations learned from the 4 mel-spectrograms. A distinct difference can be visualized from feature dimension 1700. The last 2048 dimensions are representations learned from the -60 dB and the -75 dB mel-spectrograms, these are the dimensions where the feature means deviate the most. Quite interestingly, the -60 dB and -75 dB spectrogram features yield better results compared to the others. After training the model with preprocessed signals (resampled to 1000 Hz and band-pass filtered between 20-400 Hz), the representation differences in the mean reduced for certain dimensions. This could mean that the corresponding dimensions represent information from the higher end of the frequency spectrum. Another observation experienced during experimentation was the Normal Recall vs Mild/Severe Recall trade-off. While training an end to end LP-tConv model, we have seen a divergent behavior between the normal and mild/severe recall metrics (Fig. \ref{recall}) which persisted even when the percentage of \emph{Normal} recordings were more than \emph{Mild} recordings. 

\section{Conclusions}
In this work, we have presented an ensemble of classifiers for automatically detecting abnormal heart sounds of different severity levels for the INTERSPEECH 2018 ComParE Heart Beats Sub-Challenge. The primary framework was based on transfer learning of parameters from a 1D-CNN model pre-trained on the Physionet HS Classification dataset. We have also deployed unsupervised feature representation learning from mel-spectrograms using a deep autoencoder based architecture. Finally, we have also implemented a segment-level feature based system using the ComParE feature set and an SVM classifier. The final hierarchical ensemble of the systems provided with a UAR of 57.9\% on the development dataset and 42.1\% on the test dataset.

\section{Acknowledgement}
The Titan X Pascal used for this research was donated by the NVIDIA Corporation.

% , which provides an absolute improvement of 1.7\% over the best fused system provided by the organizers of the challenge organizers.
\bibliographystyle{IEEEtran}
\balance
\bibliography{mybib}
\end{document}